\documentclass[article,12pt]{elsarticle}
\usepackage[utf8]{inputenc}
\usepackage[T1]{fontenc}
\usepackage{graphicx}
\usepackage{xcolor}
\usepackage{multicol}
\usepackage{amsmath}

\graphicspath{ {Input Material/Figures/} }

\makeatletter
\def\ps@pprintTitle{%
  \let\@oddhead\@empty
  \let\@evenhead\@empty
  \def\@oddfoot{\reset@font\hfil\thepage\hfil}
  \let\@evenfoot\@oddfoot
}
\makeatother

\begin{document}

\begin{frontmatter}

\title{Beyond Traditional DoE: Deep Reinforcement Learning for Optimizing Experiments in Model Identification of Battery Dynamics}

\affiliation[1]{Eatron Technologies Ltd., Research and Development, Warwick, CV34 6UW, United Kingdom}
\affiliation[2]{Istanbul Technical University Artificial Intelligence and Data Science Application and Research Center, Istanbul, Turkey}

\affiliation[3]{ure@itu.edu.tr}

\author[1]{Gokhan Budan}
\author[1]{Francesca Damiani}
\author[1]{Can Kurtulus}
\author[2,3]{Nazim Kemal Ure}

\begin{abstract}
Model identification of battery dynamics is a central problem in energy research; many energy management systems and design processes rely on accurate battery models for efficiency optimization. The standard methodology for battery modelling is traditional design of experiments (DoE), where the battery dynamics are excited with many different current profiles and the measured outputs are used to estimate the system dynamics. However, although it is possible to obtain useful models with the traditional approach, the process is time consuming and expensive because of the need to sweep many different current-profile configurations. In the present work, a novel DoE approach is developed based on deep reinforcement learning, which alters the configuration of the experiments on the fly based on the statistics of past experiments. Instead of sticking to a library of predefined current profiles, the proposed approach modifies the current profiles dynamically by updating the output space covered by past measurements, hence only the current profiles that are informative for future experiments are applied. Simulations and real experiments are used to show that the proposed approach gives models that are as accurate as those obtained with traditional DoE but by using 85\% less resources.
\end{abstract}


%%\item Research highlight 1
%%\item Research highlight 2
%%\end{highlights}%


\end{frontmatter}

\section{Introduction}
Modelling is an integral part of designing complex engineering systems \cite{wellstead1979introduction}; many successful applications are built upon high-fidelity models that can be analyzed via simulations to optimize the design process. The main motivation for modelling stems from the fact that a significant portion of engineering design principles that guarantee robustness, stability and high accuracy are model-based approaches \cite{jensen2011model}. The availability of a model enables quantitative analysis, optimization and simulation-based verification, which are the cornerstones of modern engineering design processes. Therefore, it is unsurprising that a significant amount of effort has been spent on developing experimental guidelines for the parameter estimation of models \cite{ljung1999system}. Modern models usually have many parameters, which makes the estimation process highly challenging. In addition, because of the dynamics of the underlying process, a single experiment is usually insufficient to cover the whole parameter space, therefore multiple experiments are required to identify the parameters of large-scale models. To this end, design of experiments (DoE), a branch of applied statistics, has been used widely across many sectors including (but not limited to) medicine, engineering, biochemistry, materials, physics and chemistry \cite{durakovic2017design}. The main objective of DoE is to develop a well-structured plan for optimizing the sequence of the experiments to maximize efficiency by obtaining high-accuracy parameter estimations in minimum time \cite{box1978statistics}.

In recent years, as the timelines for engineering projects have become more strict and the complexity of models has increased, the need to improve the DoE process has become a more prominent objective \cite{maxwell2017designing}. A well-implemented DoE plan can save significant time and cost, especially for industries where high-accuracy models are needed. For instance, the electrification of mobility and energy systems is facing significantly increased demands in terms of improved battery life and performance, thereby increasing efforts to develop new batteries and models \cite{hu2011electro,liu2012integrated,he2012comparison}. Furthermore, the application of DoE in engineering accounts for 11\% of all DoE applications, with engineering being second in popularity only to medicine \cite{durakovic2017design}. Therefore, optimizing DoE plans in the energy sector is now more critical than ever to minimize the cost of experiments and deliver products on time.

That said, although classical DoE approaches aim to explore the experiment space in an efficient manner, they do not always scale well when the numbers of parameters that must be estimated from the experiments are large and the underlying systems have complex dynamics. In energy applications, many underlying dynamical systems, such as batteries, have higher-order nonlinear dynamics, and accurate modelling of such systems requires concurrent estimation of many different parameters \cite{hu2012comparative, waag2014critical, fotouhi2015electric, young2013electric}. Although efforts have been made to advance standard DoE approaches for battery model identification \cite{makela2017experimental,su2016identifying}, optimizing DoE for realistic energy applications is still largely an open problem.

Recently, machine learning (ML) methods have begun to obtain unprecedented success across several disciplines \cite{lecun2015deep}, showing that virtually any data-driven application might benefit from using ML. In particular, ML is becoming popular in battery applications \cite{aykol2020machine}, such as for estimating remaining useful life \cite{ren2018remaining, mansouri2017remaining,wang2021critical,cheng2021remaining}, optimizing charging protocols \cite{attia2020closed, abdullah2021reinforcement} and prediction of capacity degradation\cite{severson2019data,li2022towards,tran2022python}, state of health \cite{pan2018novel} and state of charge \cite{hu2015advanced, zahid2018state, chemali2018state,babaeiyazdi2021state,chandran2021state}, fault identification/diagnostics~\cite{samanta2021machine,xue2021fault}, as well as DoE strategies in other domains such as engine calibration \cite{wong2018efficient}, bio-process improvement \cite{rodriguez2021design} and materials design \cite{wen2019machine}. Encouraged by these successful ML applications in battery dynamics, the central objective herein is to pursue ML methods for optimizing the DoE process for data-efficient battery-model parameter identification. 

\subsection{Contributions}

The main contribution of this work is a novel DoE framework based on deep reinforcement learning (RL) \cite{mnih2015human} methodologies extended to optimize the sequence of experiments in model identification of battery dynamics. We validate our approach in both computer simulations and laboratory tests on batteries. Our RL-based DoE method optimizes the combination and sequence of charge and discharge current profiles to collect battery-pack behaviour data in significantly less time than with traditional DoE methods. We use this DoE method to extract as much information as possible from the battery pack under test by respecting certain safety and feasibility criteria with an objective of planning and performing experiments without requiring any human resources. Specifically, the proposed DoE method reduces the experimental time by 85\% compared to the traditional DoE method without trading off the quality of the experimental data. The measure of effectiveness (MoE) for the quality of the experimental data is quantified with the performance metrics in the time and frequency domains. Collecting cell voltage readings from a battery pack in the high (i.e. between 20~Hz and 1~kHz), medium (i.e. 1--20~Hz) and low (i.e. less than 1~Hz) frequency ranges helps understand the resistive, electrochemical-kinetic and diffusion qualities, respectively \cite{alavi2015time}. Therefore, in the frequency domain, a uniform distribution of the aforementioned frequency ranges for the collected cell voltage measurements from the experiments is defined as one of the MoEs. Similarly, in the time domain, a uniform distribution of the cell voltages, the current applied to the battery pack, and the duration of the experiment are used as MoEs. Using simulations and real experiments, we show that our approach can obtain a model as accurate as the ones obtained with traditional DoE, using 85\% less amount of resources.

\section{Methodology}

We seek to create a single agent that can generate DoE profiles for various ranges of power, energy and chemistry of battery cells. To this end, we developed a novel agent that combines a deep neural network (see subsection "Model Architecture"), representation of the battery state, the DoE profile action space and a reward mechanism based on the TD3 RL algorithm \cite{fujimoto2018addressing} (see subsection "Algorithm"). The agent requires an observation space---the input data as a representation of the current environment---periodically to make sense of the environment (battery pack) in which it is operating. After the input data are fed through several hierarchical layers of the neural network, the agent builds up an abstract representation of the current battery-pack state and then produces a single continuous-value output to reach the design objectives by maximizing the notion of cumulative future reward. In the context of RL, a reward can be considered as a bridge that connects the motivations of an agent with that of the objective. 

\subsection{Observation Space and Reward Mechanism}

\begin{figure}[t!]
    \centering
    \includegraphics[width=\linewidth,keepaspectratio]{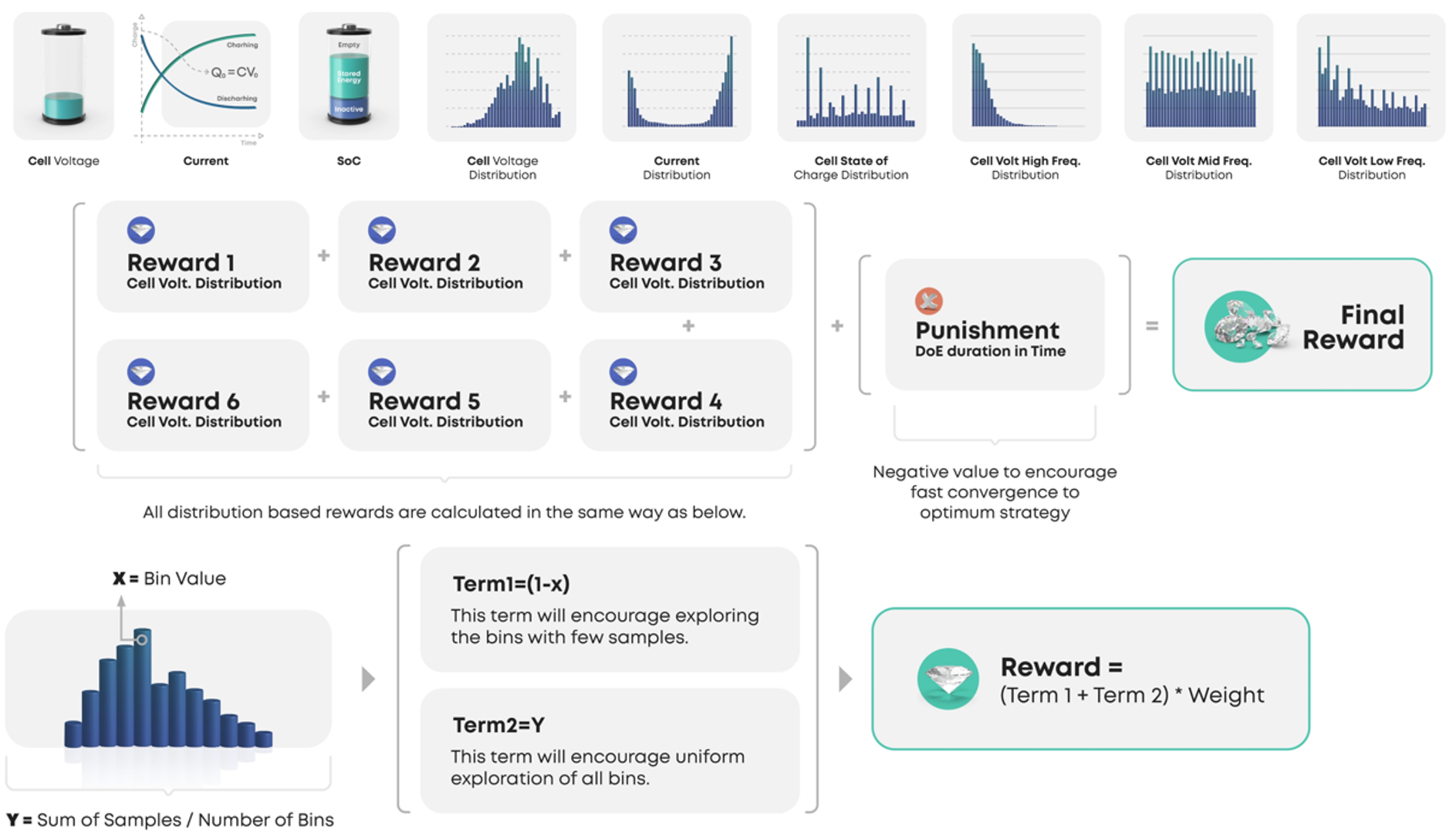}
    \caption{Reinforcement learning (RL) observation vector and reward mechanism. The observation vector shown contains key information about the state of the battery and the history of actions taken so far in order to direct future action selection. The agent must develop a long-term strategy to obtain various dynamic responses from the battery pack. The cell voltages, electrical current and state of charge (SoC) are scalar values that inform the artificial intelligence (AI) about the battery state, whereas the rest of the observation vector comprises a distribution of values that inform the AI about the history of actions and the explored dynamic battery region. The reward value guides the AI towards extracting more and diverse information from the battery pack under test. Time- and frequency-domain-based rewards are used, and the agent is encouraged to complete the whole design of experiments (DoE) task in the shortest time possible by penalizing longer experiments.}
    \label{fig:RL State Space}
\end{figure}

The observation space shown in Fig.~\ref{fig:RL State Space} contains some key information about the state of the battery pack and the history of actions taken so far to decide on the future actions towards achieving more rewards. The observation space is constructed to help the agent execute the long-term strategy for collecting various dynamic responses from the battery pack. The cell voltages, charge/discharge currents and SoC in the observation space are scalar values that reflect the battery state, whereas the distributions of values in the observation space capture the history of battery operation including information about the explored dynamical regions of the battery pack. The action space is much simpler than the observation space because it comprises only one scalar value, which is the charge/discharge current to be applied to the battery pack in the next step. This output from the neural network is between $-1$ and 1, which is then scaled accordingly to the correct ampere value depending on the battery-pack manufacturer guidelines. We found that implementing a battery-pack-dependent scaling of the agent actions ensure that (i) the DoE method can be generalized to other types of battery cells because the essential behaviour and dynamic response of battery cells show similar signatures \cite{mauracher1997dynamic} and (ii) it is feasible to apply the generated actions with the battery cycler equipment in the laboratory because that equipment is limited in how fast it can respond to and update the DoE control parameters. The reward mechanism is another crucial consideration; it essentially motivates the agent to select or ignore certain actions in a given battery-pack state. As shown in Fig.~\ref{fig:RL State Space}, the reward mechanism is geared towards making the agent extract more-diverse information from the battery pack. In practice, a reward is a scalar value that is the weighted sum of individual reward factors to encourage or discourage the taken action. As mentioned previously, time- and frequency-domain-based MoEs are used in the reward mechanism, and the agent is encouraged to complete the DoE task in the shortest time possible by giving a punishment in the form of a negative reward value at each timestep at which DoE is executed.

\subsection{Stages of AI Based DOE}

\begin{figure}[t!]
    \centering
    \includegraphics[width=\linewidth,keepaspectratio]{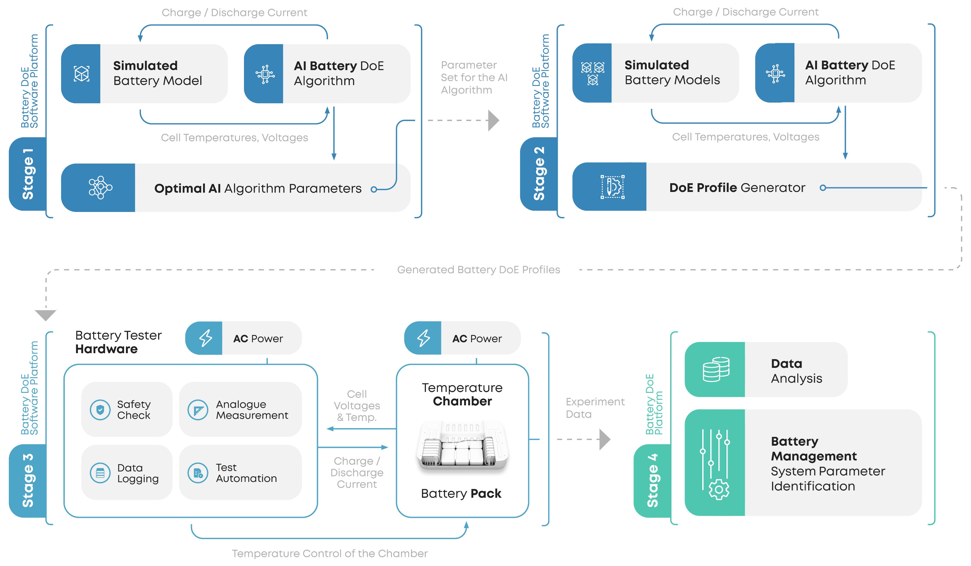}
    \caption{AI battery DoE stages. In stage~1, the AI algorithm is trained using multiple simulated battery models so that the agent observes all possible states that a battery pack can visit. In stage~2, the AI algorithm interacts with a single model of the battery pack to generate the DoE electric-current profiles. The stage-2 output comprises the battery DoE electric-current profiles, which essentially determine the length of stage-3 testing in addition to the magnitude and frequency of the electric currents for each timestep to be applied to the battery pack under test.}
    \label{fig:Stages of DoE}
\end{figure}

As shown schematically in Fig.~\ref{fig:Stages of DoE}, our method has four key stages. In stage~1, the agent is trained using multiple simulated battery models running in parallel, and these are initialized with various ranges of SoC and temperature to accelerate the training process (see subsection "Training details"). The training could also be done with a physical battery pack in a laboratory environment, but data collection and real-time interaction with a physical battery are slow and costly, so the simulation method was chosen for training the agent. At the end of stage~1, we obtain the parameter set for the neural network of the agent that is optimized to generate battery DoE profiles. In stage~2, the trained agent interacts with a single battery model in simulation to generate the target DoE profile. In stage~3, the generated battery DoE profile is applied to a physical battery pack in a laboratory environment to collect experimental data. In stage~4, the parameter identification for the equivalent-circuit model (ECM) of the battery pack is done based on the collected data such as voltage, current and temperature. The main outcome of this stage-based process is the parameter set for the analytical battery model that essentially describes the expected dynamical behaviour under all operating conditions. 

\subsection{Experimental Details}

\begin{figure*}[t!]
    \centering
    \includegraphics[width=\textwidth,keepaspectratio]{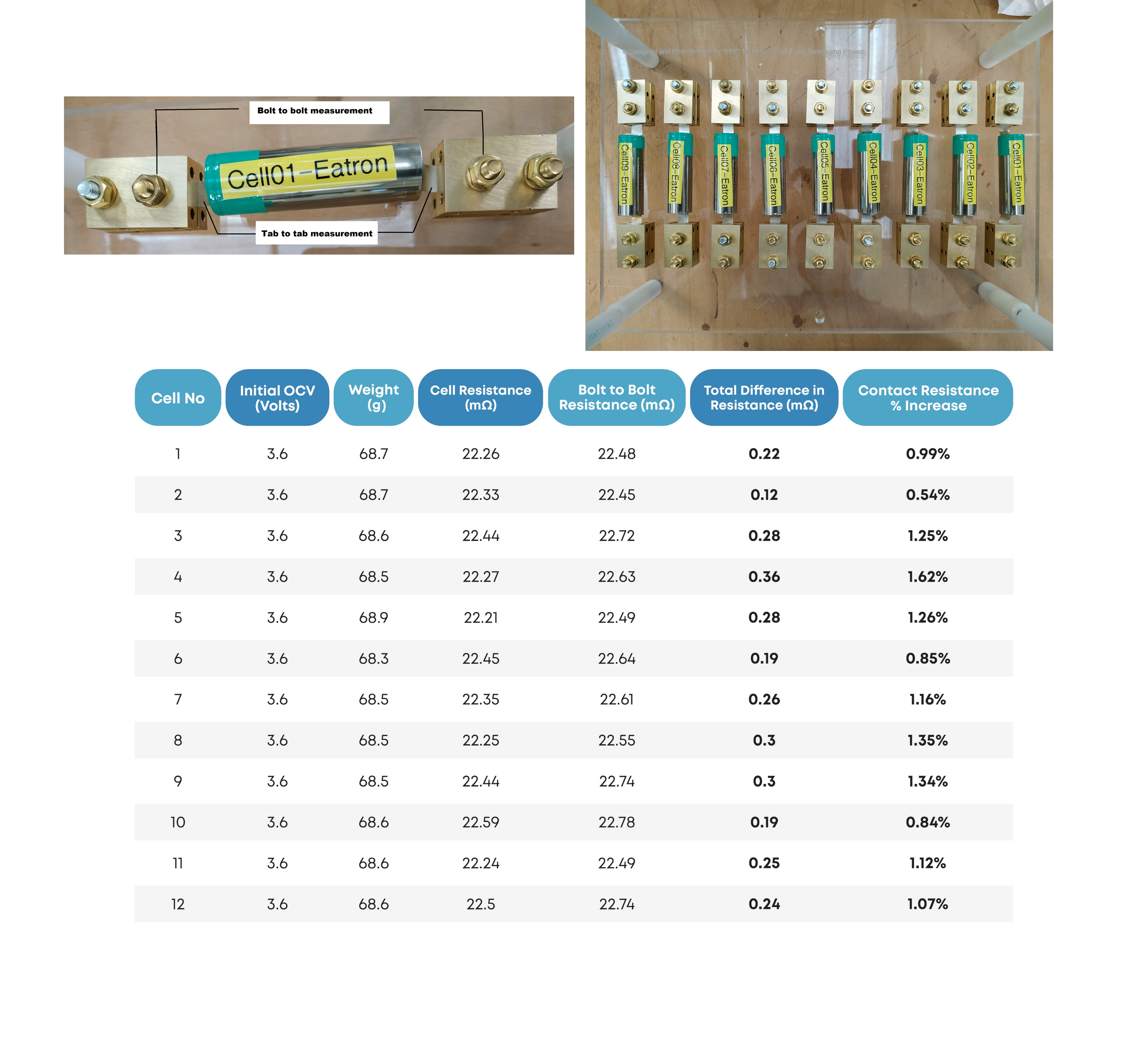}
    \caption{\textbf{Rig design for cell characterization and how the cells are connected with the cycle.} The table gives some values from initial tests (e.g., bolt-to-bolt resistance).}
    \label{fig:Figure_7}
\end{figure*}

\begin{figure}[!]
    \centering
    \includegraphics[scale=0.5]{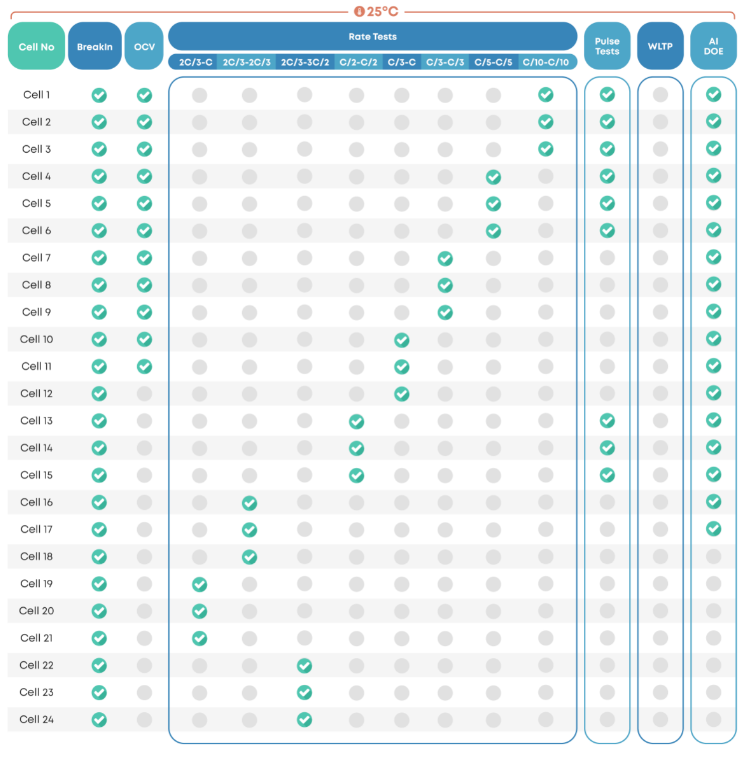}
    \caption{\textbf{During the experimental tests, 24 cells were used.} Each cell experienced a break-in cycle at 25$^{\circ}$C. Break-in cycle is usually performed at the beginning of an experiment and it involves charging and discharging cells in sequence to excite slowly unused sections of a battery pack; it is used to precondition the cell and is not part of the identification procedure. After that, OCV rate tests (involving a sequential charge--discharge cycle at really low C rates) at 25$^{\circ}$C were performed on 12 cells to characterize the SoC--OCV relationship. Later, rate tests were performed on eight groups of three cells each. These tests involved charging and discharging the cell at a constant C rate. As given in the table, the C rate used in the test was different for each group. To acquire information about the dynamic part of the cell (i.e. ohmic resistance), pulse tests were performed on six different cells, involving a series of pulses of charge and discharge for a fixed duration and amplitude followed by a rest period of the same duration. During this test, the amplitude and duration were varied in a predefined way to ex the cell with different frequencies, and the test was repeated at different SoCs. A Worldwide Harmonised Light Vehicles Test Procedure (WLTP) \cite{pavlovic2018much} was then performed on three different cells to be used as a test profile describing a normal operating scenario for the battery. Finally, the AI DoE generated by the deep RL algorithm was performed for 17 cells.}
    \label{fig:Figure_8}
\end{figure}

We generated two DoE profiles based on the traditional method and the proposed artificial intelligence (AI) method and applied them to a physical 21700 (LGM50) lithium-ion battery pack in a laboratory environment and to the battery-pack model in simulation for battery parameter identification (see subsection "Parameter identification" ). The traditional DoE involved both constant and dynamic current profiles to capture both slow and fast battery dynamics covering the whole SoC range (see Fig.~\ref{fig:Figure_7} and Fig.~\ref{fig:Figure_8}). The total experimental time was 170~h with the traditional DoE method but only 26~h with the proposed AI method, a significant reduction of 85\%. Battery parameter identification was performed with the four collected sets of data (the AI and traditional DoE methods in computer simulation and laboratory experiments). The parameterized battery model was then tested with the Federal Test Procedure 75 (FTP75) \cite{epa_2020} drive-cycle profile, which had not been used in any of the previous experiments. In the rest of the paper, we provide details on battery model, parameter identification, traditional DOE method and the training process of the AI DOE model.

In this work, we used 24 cylindrical high-energy 21700 (LGM50) lithium-ion cells that had a nominal capacity of 5~Ah and a nominal voltage of 3.6~V. The height, diameter and weight of the cells were 70.80~mm, 21.44~mm and 70.0~g, respectively. The charge and discharge currents are defined in units of C rate; for instance, 1C refers to 5~A, which is the current required to fully charge or discharge a cell in 1~h. The ESPEC thermal chamber used during the cell and module testing was rated at level~6 on the EUCAR hazard scale, which refers to the ability of the temperature chamber to mitigate the danger from a potential hazard associated with a lithium-ion battery. The calibration of the battery cycler (Digatron) was validated before the start of the experimental work. For the temperature measurements, we used K-type thermocouples that were accurate to 1$^{\circ}$C. 

All of the cells were subjected initially to a set of characterization tests to obtain the reference performance metrics at different operating conditions, during which current, voltage and temperature were measured. The characterization tests and the conditions are summarized in Fig.~\ref{fig:Figure_8}. With the data obtained from these initial tests, the cell capacities at different C rates and ambient temperatures were determined, the pulse power capabilities at different SoCs and ambient temperatures were obtained, and the relationship between open-circuit voltage (OCV) and SoC was found.

%Following the cell characterisation tests, the module level testing (12S1P) was carried out. Voltage and temperature values were measured from each cell using a \colorbox{yellow}{Hioki/Pico} unit. Firstly, the module is charged from \colorbox{yellow}{x\% to x\%} SoC with a 0.3C CC-CV charging step to \colorbox{yellow}{X V} and a current cutoff of \colorbox{yellow}{C/X}. Secondly, cell balancing was conducted using a programmable power supply. Thirdly, a \colorbox{yellow}{X} period of rest was applied. Fourthly, the module was discharged with a 1 CC-CV to \colorbox{yellow}{X V} and a current cutoff of \colorbox{yellow}{C/X}.%

\subsection{Battery Model}
ECMs are used frequently to understand the behaviour of a cell in response to a stimulus, which is a current waveform \cite{huria2012high}. The battery model used in the simulations was an ECM parameterized with data from a lithium polymer (LiPo) pouch cell with a nominal capacity of 15~Ah. These data are provided as a supplement to Gregory Plett's "Battery Management Systems" book.  \cite{plett2015battery} .

The ECM comprises the following.
\begin{itemize}
\item An ideal voltage source as the open-circuit voltage. This is implemented as a function of the SoC and temperature through a two-dimensional look-up table. 
\item An equivalent series resistance that models the instantaneous drop in cell voltage under a load. This is the pure resistance effect of high-frequency behaviour and is modelled as a function of SoC, temperature and current through a three-dimensional lookup table. The dependency on the current is needed to simulate the Bulter--Volmer equation, which describes how the electrical current on an electrode depends on the electrode potential \cite{buller2003impedance}.
\item The non-instantaneous drop due to low-frequency behaviour is modelled with three RC pairs: one models the drop caused by the combination of double-layer capacitance and charge-transfer resistance \cite{jossen2006fundamentals}, while the other two model the drop due to diffusion. All these parameters are implemented as a function of SoC and temperature.
\end{itemize}
The SoC is calculated using Ah counting with the battery capacity modelled as a function of temperature with a one-dimensional lookup table.

\subsection{Traditional DOE Method}

Traditional DoE is a test procedure performed at cell level to calibrate the ECM via test measurements. 

The DoE needs to vary depending on the final application of the ECM (i.e. for use in an automotive battery management system or a stationary application), and designed that the amplitude of the current is defined without violating the constraints of the cell manufacturer. It is also important that the DoE is specified in order to exercise the full ranges of SoC, temperature and C rate; in this way, the parameter set can describe the behaviour of the cell in different scenarios. The present scope is to investigate the procedure at 25$^{\circ}$C, but future work might involve investigating the proposed procedure at different temperatures. Traditional DoE uses three different procedures: (i) constant-current profiles, (ii) pulse profiles and (iii) drive cycles. The aim of a constant-current test is to capture the current dependency; it is carried out while charging and discharging the cell completely at different rates, and during the slow dynamics it is also possible to extract the long time constants of diffusion. A pulse procedure is a more dynamic test in which pulses are applied over the whole SoC range with different durations and amplitudes to obtain information about the faster dynamics through the voltage drop; it is also possible to obtain information about the slower dynamics during the transients. Finally, a drive cycle is used mainly to validate the parameter set under conditions similar to those in real-world driving scenarios.

\subsection{Parameter Identification}

\begin{figure*}[t!]
    \centering
    \includegraphics[scale=0.4]{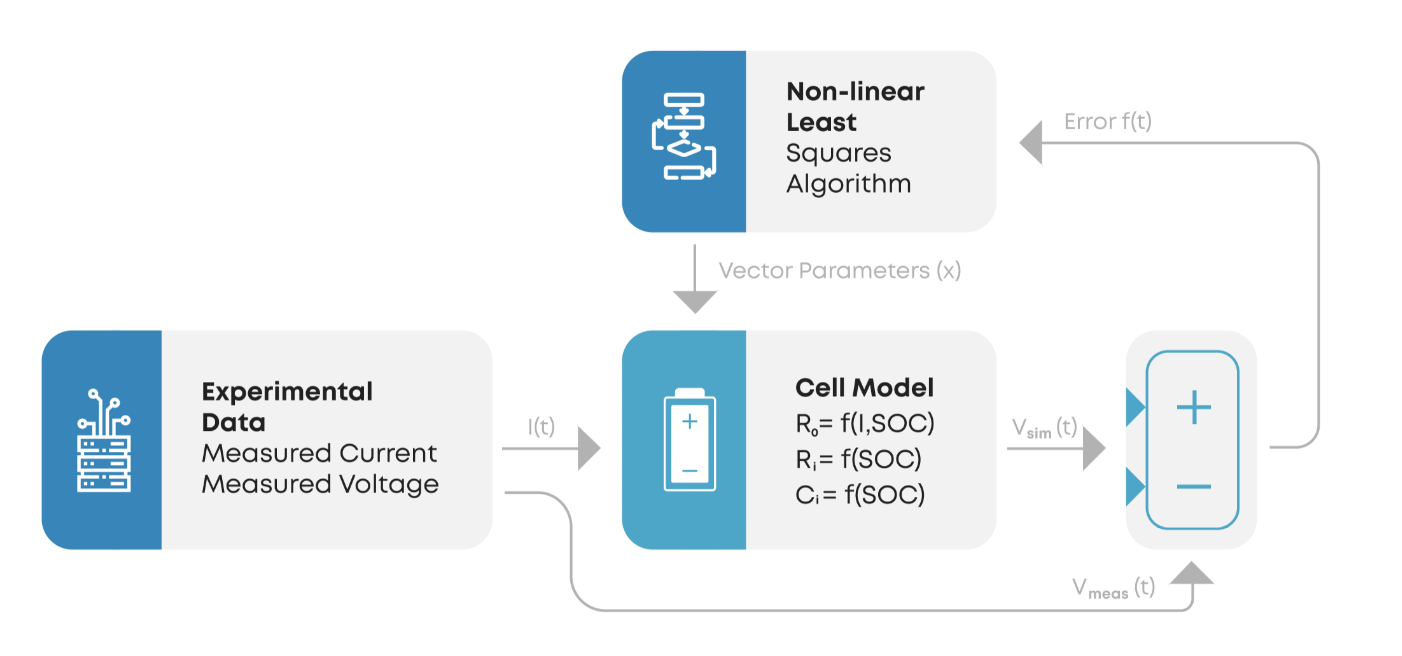}
    \caption{\textbf{Approach used to identify model parameters.} The measurements are fed into the cell model and the error is evaluated until the nonlinear least-squares algorithm converges.}
    \label{fig:Figure_5}
\end{figure*}

To compare the quality of the proposed AI-based DoE with that of the traditional one, experimental data from both methods are used to fit a discrete-time cell model. The cell model used during identification is the same as that used to train the AI algorithm except for the number of RC pairs. The cell model used for identification has two RC pairs, therefore the parameters to be identified are the ohmic resistance $R_o = f_1(SoC,I)$ and the two RC pairs $R_i = f_2(SoC)$ and $C_i = f_3(SoC)$. The aim is to find the parameter values that fit the model best to the measurements taken from the real battery; to do so, a nonlinear least-squares identification algorithm is used \cite{xiong2011modeling}.

Given the vector of parameters $x$ and $f_i(x),\,i=1,\ldots ,n$, which are the differences between the simulated and measured terminal voltage for all the measurement points, the optimum parameter $\hat{x}$ is determined as 
\begin{equation} \label{eq1}
\begin{split}
\hat{x} = \min_{x} \Big|\Big|f(x)\Big|\Big|^2_2 = \\
\min_{x} \Big(f_1(x)^2 + f_2(x)^2 + ... + f_n(x)^2\Big), 
\end{split}
\end{equation}
where
\begin{equation} \label{eq2}
f(x) = \left[ \begin{array}{c}
f_1(x) \\
f_2(x) \\
\vdots \\
f_n(x) \end{array} \right].
\end{equation}

The identification is performed using the data from both the traditional and AI DoE, giving two sets of parameters that fit the data ($\hat{x}_{Trad}$ and $\hat{x}_{AI}$, respectively). The cell model is then evaluated using experimental data that were not used during the identification process, using the two different sets of parameters. The terminal voltages are then compared with the true one, and the results are presented in the Results section. Overall parameter identification process is displayed in Fig. \ref{fig:Figure_5}. 
\subsection{Algorithm}
We consider an agent that is tasked with executing a sequence of actions (i.e. determining charge and discharge current values) while interacting with a stochastic environment (i.e. the battery pack under test) based on the observations made and the rewards obtained. At each timestep, which is 1~ms, the agent outputs a single scalar value between $-1$ and 1 that corresponds to a charge current if above zero and a discharge current if below zero. This output value is then scaled to an actual current value within the maximum allowed charge and discharge currents (i.e. 120~A and $-60$~A, respectively) based on the cell datasheet and the battery-pack setup. It is important to note that the reward received is highly dependent on the whole previous sequence of charge and discharge current values and observations because any useful feedback from an electrochemical environment such as a battery pack can be received only after thousands of timesteps have elapsed.

The goal of the agent then becomes selecting charge and discharge currents that maximize the future rewards. To this end, we used the TD3 algorithm, which builds on the DDPG algorithm \cite{lillicrap2015continuous}. 

\subsection{Model architecture}
The model architecture is based on the actor--critic method \cite{sutton2018reinforcement}, in which the critic updates the parameters of the value function and the actor updates those of the policy. The actor--critic method is deployed under the TD3 algorithm, and there are six identical neural networks: two critics, two critic targets, one actor and one actor target. The neural-network architecture into which the observation space shown schematically in Fig.~\ref{fig:RL State Space} is fed is as follows. The input layer receives an array of 43 observation-space values that are normalized in the range $[0,1]$. This is followed by four fully connected hidden layers comprising 128, 128, 64 and 64 rectifier units, respectively. Each hidden layer is followed by a dropout layer with a fixed dropout probability of 0.2, 0.2, 0.1 and 0.1, respectively. The output layer is a fully connected layer with a single output and tanh activation to produce a value in the range $[-1,1]$. This produced output value is then scaled according to the datasheet of the battery under test to obtain the charge or discharge current to be applied at that particular timestep.

\subsection{Training Details}
Here, we give a step-by-step explanation of the training procedure and present the RL algorithm. Forward pass is the process of obtaining the output data after the battery observation vector is given as an input to the neural network, and back propagation is the process of neural-network parameter updating via the Adam optimizer \cite{kingma2014adam}. The training steps are as follows. 1) The replay buffer is initialized; the TD3 algorithm uses a replay buffer to store past transitions when executing the policy. Each transition can be represented as a tuple in the form of $(s,s',a,r)$, where $s$ and $s'$ are the current and next observations, respectively, $a$ is the selected action and $r$ is the reward value. During training, the transition tuples in the replay buffer are queried to replay the agent's experience in a shuffled way to reduce correlation between samples. 2) Next, the neural networks are built using the TensorFlow open-source machine-learning library \cite{tensorflow2015-whitepaper}. There are six neural networks in total, each with exactly the same numbers of layers and neurons, one each for the actor and actor target models and two each for the critic and critic-target models. The actor networks learn the policy $\pi(s|a)$ while the critic ones learn $Q_{\pi}(s,a)$. The rationale behind having target networks for both actor and critics is to be more conservative when updating the neural-network parameters; in other words, the target network parameters are constrained to change at a slower rate that is determined by Polyak averaging \cite{fujimoto2018addressing}:

\begin{equation} \label{eq3}
\Phi_{target} = \rho\Phi_{target} + (1-\rho)\Phi,
\end{equation}

where $\rho$ is in the range $[0,1]$ and determines the rate of change in the target network parameter set. 3) The agent starts taking actions according to the initial policy in the environment. Forward pass is repeated until the replay buffer is full, after which a batch-size number of transitions are sampled from the buffer. For each transition in the sampled batch, the actor target produces the next action $a'$, and the pair $(s',a')$ is given as an input to the two critic targets. The critic targets then return the values of the state-action pair, $Q_{tar1}(s',a')$ and $Q_{tar2}(s',a')$, independent from each other. The final critic-target $Q$ value is obtained by

\begin{equation} \label{eq4}
Q_{tar} = r + \gamma(min(Q_{tar1},Q_{tar2})),
\end{equation}

where $\gamma$ is the discount factor in the range $[0,1]$. Taking the minimum of the two $Q$ values has been found to stabilize the optimization process \cite{fujimoto2018addressing} because optimistic $Q$-value estimates are avoided by ignoring the higher $Q_tar$ value. Following this, the two critic networks take the $(s,a)$ pair as an input and produce $Q_1(s,a)$ and $Q_2(s,a)$ to compute the final critic loss:

\begin{equation} \label{eq5}
Loss_{critic} = \mathit{MSE}(Q_{1}(s,a),Q_{tar}) + \mathit{MSE}(Q_{2}(s,a),Q_{tar}),
\end{equation}

where $MSE$ refers to the mean-squared error loss. This loss value is used during backpropagation to update the critic network parameters. This training step in which the agent tries to reduce the critic loss is called the Q-learning step, the aim of which is to find the optimal parameter set for the critic networks. 4) The next step is the policy-learning step, the aim of which is to find the optimal parameter set for the actor network to maximize the expected return. The $Q$ value from a critic is correlated with the expected return, meaning that as the $Q$ value increases, the expected return moves to being optimal. In this case, the loss for the policy learning is the mean of the $Q$ values from the critics:

\begin{equation} \label{eq6}
Loss_{actor} = -\frac{(Q_{1}(s,a) + Q_{2}(s,a))}{2}.
\end{equation}

During backpropagation, gradient ascent is used by differentiating the actor loss with respect to the actor network parameters in the direction that maximizes the expected return. The important point to note here is that policy learning is done every other step whereas Q-learning in step~3 is done every step. If the Q-learning is poor, then the policy becomes poor as well and can cause divergence of the loss moving towards minima. This is why Q-learning is done at double the rate of policy learning to increase the performance of convergence to the optimal parameter set. 5) The final step in the training cycle is to update the target network parameters of the actor and the critics, which has not been done thus far. The target network parameters are updated, and this essentially copies the weights of the actor and critic networks with Polyak averaging into the target networks. As with the policy learning, the target-network updates are done every other step to improve the training performance stability. Steps~3--5 are repeated until a training stop condition is met. As the agent continues to operate in the environment, the replay buffer is overwritten with the new transitions starting from the oldest entry.

%\subection{Extended Data}

% \begin{figure*}[ht]
%     \centering
%     \includegraphics[scale=0.5]{Figure_6a.png}
%     \label{fig:Figure_6a}
% \end{figure*}

\section{Results}

\begin{figure*}[t!]
    \centering
    \includegraphics[width=\textwidth,keepaspectratio]{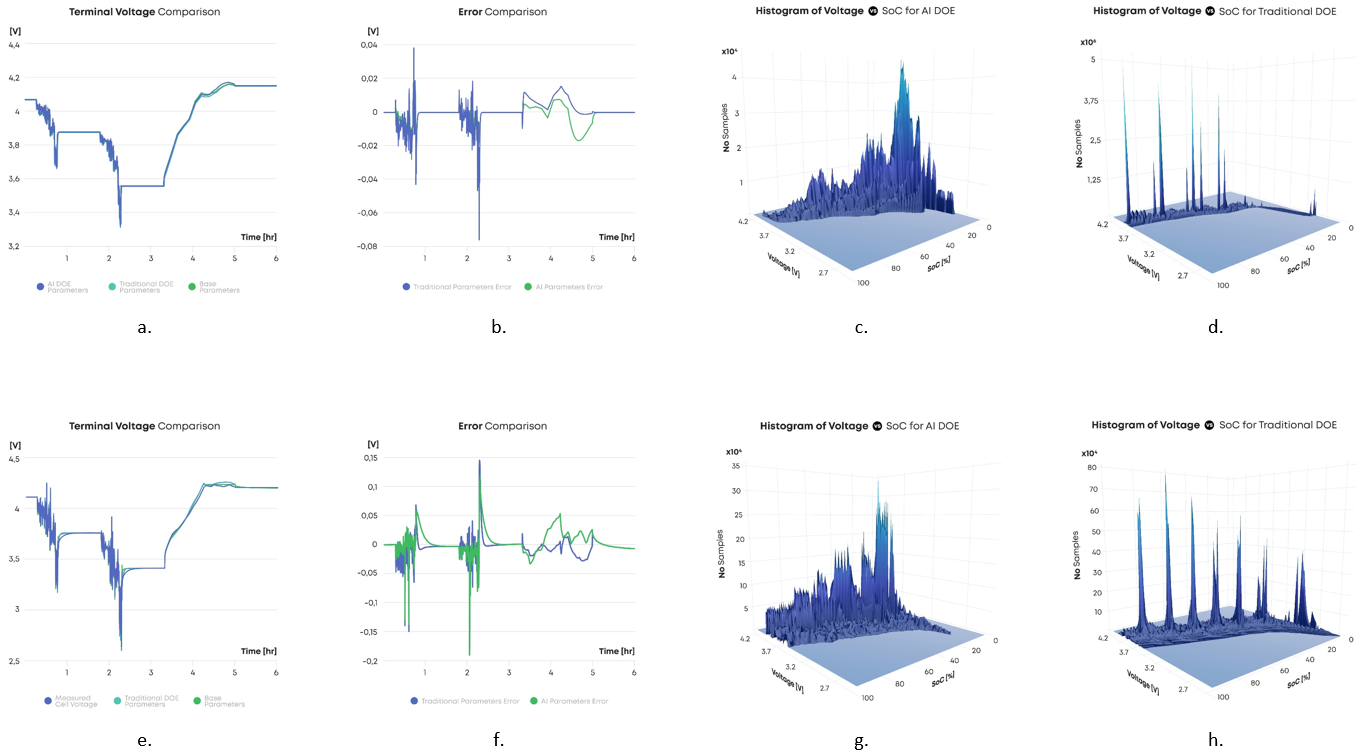}
    \caption{\textbf{a--d, Computer simulation results: a}, comparison of terminal voltage among models parameterized with base data set and those of the AI and traditional methods; \textbf{b}, comparison of errors in terminal voltage when referenced against base model; \textbf{c,~d}, histogram of terminal voltage and battery SoC against number of samples using (\textbf{c}) AI and (\textbf{d}) traditional methods. \textbf{e--h}, As \textbf{a--d} but for laboratory experiments.}
    \label{fig:Results1}
\end{figure*}

\begin{figure}[t!]
    \centering
    \includegraphics[width=\textwidth,keepaspectratio]{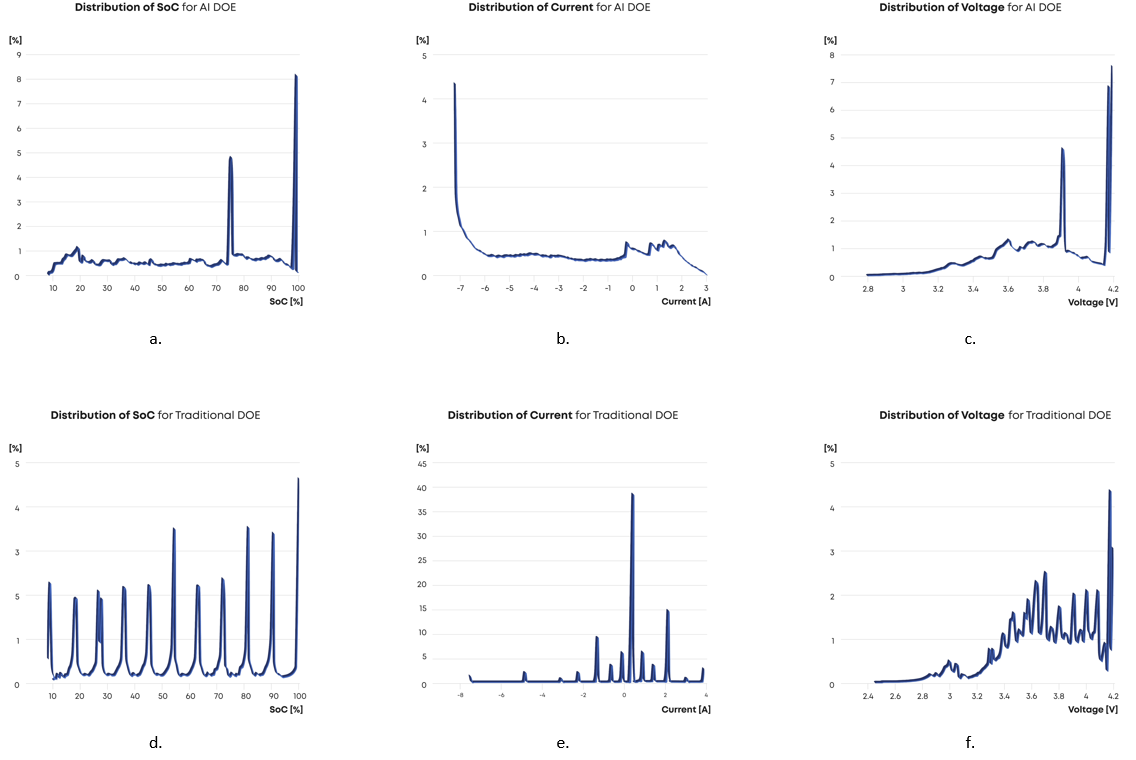}
    \caption{\textbf{a--c}, Distributions of battery SoC, charge/discharge current and terminal voltage against number of samples obtained in laboratory experiments using AI method. \textbf{d--f}, As \textbf{a--c} but using traditional method.}
    \label{fig:Results2}
\end{figure}

\begin{figure*}[ht]
    \centering
    \begin{multicols}{2}
     \includegraphics[scale=0.45]{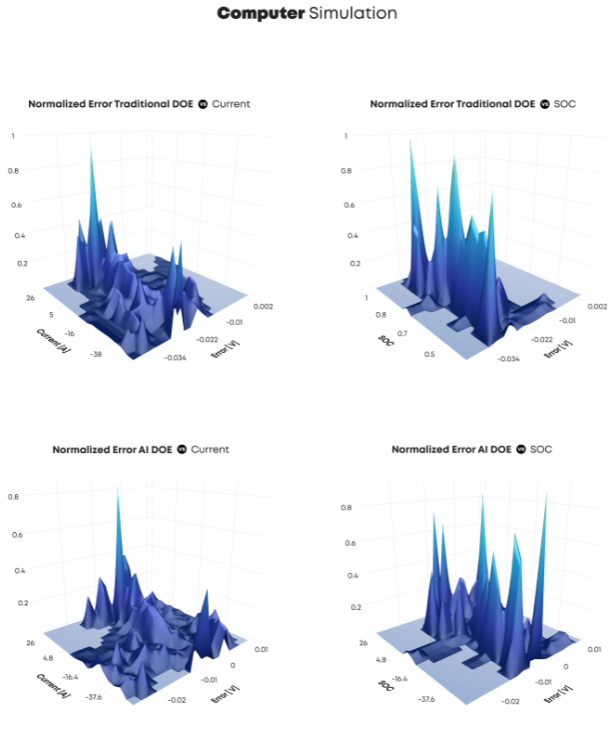}\par
    \includegraphics[scale=0.45]{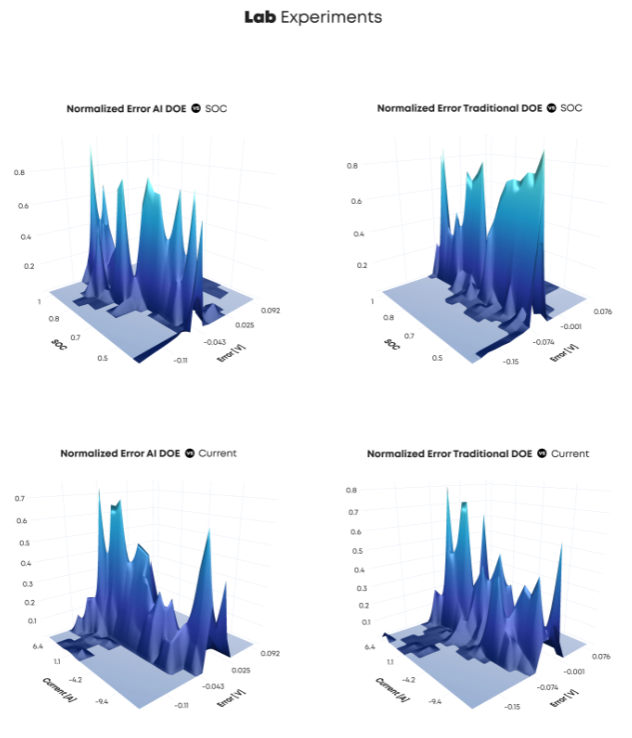}\par
    \end{multicols}
   \caption{\textbf{Resulting error distribution using parameter identification with traditional and AI DoE data.} The error is normalized with respect to SoC and current. The graphs show that the error is uniform and that all of the SoC and current regions explored during testing behaved correctly, with the error concentrated mostly around 0~V.}
    \label{fig:Figure_6b}
\end{figure*}

The validation results are presented in Fig. \ref{fig:Results1}, Fig. \ref{fig:Results2} and Fig. \ref{fig:Figure_6b}. Results show that the mean errors of the terminal voltage of the battery models parameterized with the data collected using the AI and traditional DoE methods are comparable with the mean absolute errors of 2.8~mV and 3.6~mV, respectively, in the computer simulation (Fig.~\ref{fig:Results1}b) and 10~mV and 7.3~mV, respectively, in the laboratory experiments (Fig.~\ref{fig:Results1}f). As such, the AI DoE method arguably does not trade off data quality while reducing the experimental time by 85\% compared to that with the traditional DoE method. This result also validates the ability of our method to generalize battery DoE profile generation to battery packs with capacity, voltage and current specification not seen during training.

\section{Conclusion}

In summary, we have successfully optimized the combination and sequence of input electrical-current profiles in the DoE for lithium-ion batteries by using AI techniques to reduce the time and effort of determining battery parameter sets, and we have shown that our approach gives models that are as accurate as those obtained with traditional DoE but using 85\% fewer data samples. The proposed DoE method can be extended to batteries with different chemistry (e.g.\ lithium iron phosphate (LFP), lithium nickel manganese cobalt oxide (NMC) etc.) and electrical characteristics during experiments. This research opens new opportunities for battery DoE optimization and estimation of parameters in battery-management system models. It is also important to note that DoE, in general, is a systematic method for determining the relationship between factors affecting a process and the output of that process; in other words, it is used to find cause-and-effect relationships. Therefore, our method could also be applied to other systems with multi-dimensional parameter spaces for (i) variable screening to select important factors among the many that affect the performance of a system and (ii) transfer-function identification to better understand the relationship between sets of input and output variables.

 \bibliographystyle{elsarticle-num} 
 \bibliography{references.bib}

\end{document}